\documentclass{article} 
\usepackage{iclr2017_conference,times}
\usepackage{hyperref}
\usepackage{url}
\usepackage{graphicx}
\usepackage{csquotes}

\newcommand{\Expect}{{\rm I\kern-.3em E}}

\title{Multi-Agent Cooperation \\and the Emergence of (Natural) Language}

\author{Angeliki Lazaridou$^{1}$\thanks{Work done while at Facebook AI Research.} , Alexander Peysakhovich$^2$, Marco Baroni$^{2,3}$\\
$^1$Google DeepMind, $^2$Facebook AI Research, $^3$University of Trento\\
\texttt{angeliki@google.com}, \texttt{\{alexpeys,mbaroni\}@fb.com}
}

\begin{document}

\maketitle

\begin{abstract}
  The current mainstream approach to train natural language systems is
  to expose them to large amounts of text. This passive learning is
  problematic if we are interested in developing \emph{interactive}
  machines, such as conversational agents. We propose a framework for
  language learning that relies on multi-agent communication. We study this learning in the context of referential games.   In these games, a sender and a
  receiver see a pair of images. 
   The sender is told one of them is the target and is allowed to send a message from a
   fixed, arbitary vocabulary to the receiver. 
   The receiver must rely on this message to identify the target. 
   Thus, the  agents develop their own language interactively out of the need to
  communicate. We show that two networks with simple
  configurations are able to learn to coordinate in the referential
  game. We further explore how to make changes to the game environment to cause the ``word meanings'' induced in
  the game to better reflect intuitive semantic properties of the images. In addition, we present a simple strategy for grounding the
  agents' code into natural language. 
  Both of these are necessary steps towards developing machines that are able to
  communicate with humans productively. %
\end{abstract}

\section{Introduction}

\begin{displayquote}
I tried to break it to him gently [...] the only way to learn an unknown language is to interact with a native speaker [...] asking questions, holding a conversation, that sort of thing [...] If you want to learn the aliens' language, someone [...] will have to talk with an alien. Recordings alone aren't sufficient.

Ted Chiang, \emph{Story of Your Life}
\end{displayquote}

One of the main aims of AI is to develop agents that can cooperate with others to achieve goals \citep{wooldridge2009introduction}. Such coordination requires communication. If the coordination partners are to include humans, the most obvious channel of communication is natural language. Thus, handling natural-language-based communication is a key step toward the development of AI that can thrive in a world populated by other agents.

Given the success of deep learning models in related domains such as image captioning or machine
translation \citep[e.g.,][]{Sutskever:etal:2014,Xu:etal:2015}, it
would seem reasonable to cast the problem of training conversational
agents as an instance of supervised learning \citep{Vinyals:Le:2015}.
However, training on ``canned''
conversations does not allow learners to experience the interactive
aspects of communication. Supervised approaches, which focus on
the structure of language, are an excellent way to learn
general statistical associations between sequences of symbols. However, they do
not capture the functional aspects of communication, i.e., 
that humans use words to coordinate with others and 
make things happen \citep{Austin:1962,clark1996using,Wittgenstein:1953}.

This paper introduces the first steps of a research program based on
\textit{multi-agent coordination communication games}. These games
place agents in simple environments where they need to develop a language to 
coordinate and earn payoffs. Importantly, the agents start as blank slates,
but, by playing a game together, they can develop and bootstrap
knowledge on top of each others, leading to the emergence of a
language.

The central problem of our program, then, is the following: How do we
design environments that foster the development of a language that is
portable to new situations and to new communication partners (in
particular humans)?

We start  from the most basic challenge of using a language
in order to \emph{refer} to things in the context of a two-agent game.
We focus on two questions. First, whether \emph{tabula rasa} agents
succeed in communication. Second, what features of the environment 
lead to the development of codes resembling human
language. 

We assess this latter question in two ways. 
First, we consider whether the agents associate general
conceptual properties, such as broad object categories (as opposed to 
low-level visual properties), to the symbols they learn to use. Second, we 
examine whether the agents' ``word usage'' is partially interpretable by 
humans in an online experiment.

Other researchers have proposed communication-based environments for
the development of coordination-capable AI. 
Work in multi-agent systems has focused on the design of
pre-programmed communication systems to solve specific tasks (e.g.,
robot soccer, \citealt{stone1998towards}).  Most related to our work, \cite{sukhbaatar2016learning} and \cite{Foerster:etal:2016}
show that neural networks can evolve communication
in the context of games without a pre-coded protocol.
We pursue the same question, but further ask how we
can change our environment to make the emergent language more interpretable.

Others (e.g., the SHRLDU program of \citealt{Winograd:1971} or the game in \citealt{Wang:etal:2016}) propose building a communicating AI by putting humans in the loop from the very beginning. This approach has benefits but faces serious scalability issues, as active human intervention is required at each step. An attractive component of our game-based paradigm is that humans may be added as players, but do not need to be there all the time.

A third branch of research focuses on ``Wizard-of-Oz'' environments,
where agents learn to play games by interacting with a complex
scripted environment~\citep{Mikolov:etal:2015a}. This approach gives
the designer tight control over the learning curriculum, but imposes a
heavy engineering burden on developers. We also stress the
importance of the environment (game setup), but we focus on simpler environments with multiple agents that force them to get smarter by bootstrapping on top of each
other. 

We leverage ideas from work in linguistics, cognitive science and game
theory on the emergence of language 
\citep{Wagner:etal:2003,Skyrms:2010,
  crawford1982strategic,crawford:1998survey}. 
Our game is a variation
of Lewis' signaling game \citep{Lewis:1969}. There is a rich
tradition of linguistic and cognitive studies using similar setups
\citep[e.g.,][]{Briscoe:2002,Cangelosi:Parisi:2002,Spike:etal:2016,Steels:Loetzsch:2012}. What
distinguishes us from this literature is our aim to, eventually,
develop practical AI. This motivates our focus on more realistic input
data (a large collection of noisy natural images) and on trying to
align the agents' language with human intuitions.

Lewis' classic games have been studied extensively in game theory under the name of ``cheap talk''. These games have been used as models to study the evolution of
language both theoretically and experimentally
~\citep{crawford:1998survey,blume:1998experimental,crawford1982strategic}. A
major question in game theory is whether equilibrium actually occurs in a game as
convergence in learning is not guaranteed
\citep{fudenberg2014recency,roth1995learning}. And, if an equilibrium is reached, which one it will be
(since they are typically not unique). This is particularly true for
cheap talk games, which exhibit Nash equilibria in which precise language emerges, others where vague language
emerges and others where no language emerges at all
\citep{crawford1982strategic}. In addition, because in these games
language has no ex-ante meaning and only emerges in the context of the
equilibrium, some of the emergent languages may not be very natural.  Our results speak to both the convergence question and the
question of what features of the game cause the appearance of
different types of languages. Thus, our results are also of interest
to game theorists.

An evolutionary perspective has recently been advocated as a way
to mitigate the data hunger of traditional supervised approaches
\citep{Goodfellow:etal:2014, Silver:etal:2016}. This research confirms that 
learning can be
bootstrapped from \emph{competition} between agents. We focus,
however, on \emph{cooperation} between agents as a way to foster
learning while reducing the need for annotated data.

\section{General Framework}
\label{sec:general-framework}

Our general framework includes K players, each parametrized by $\theta_k$, a collection of tasks/games that the players have to perform, a communication protocol $V$ that enables the players to communicate with each other, and payoffs assigned to the players as a deterministic function of a well-defined goal. In this paper we focus on a particular version of this: \textit{referential games}. These games are structured as follows.

\begin{enumerate}
\item There is a set of images represented by vectors $\lbrace i_1, \dots, i_N \rbrace$, two images are drawn at random from this set, call them $(i_L, i_R)$, one of them is chosen to be the ``target'' $t \in \lbrace L, R \rbrace$
\item There are two players, a sender and a receiver, each seeing the images - the sender receives input $\theta_S (i_L, i_R, t)$
\item There is a \textit{vocabulary} $V$ of size $K$ and the sender chooses one symbol to send to the receiver, we call this the sender's policy $s(\theta_S (i_L, i_R, t))\in V$
\item The receiver does not know the target, but sees the sender's symbol and tries to guess the target image. We call this the receiver's policy $r(i_L, i_R, s(\theta_S (i_L, i_R, t))) \in \lbrace L, R \rbrace$ 
\item If $r(i_L, i_R, s(\theta_S (i_L, i_R, t)) = t$, that is, if the receiver guesses the target, both players receive a payoff of 1 (win), otherwise they receive a payoff of 0 (lose).
\end{enumerate}

Many extensions to the basic referential game explored here are possible. There can be more images, or a more sophisticated communication protocol (e.g., communication of a sequence of symbols or multi-step communication requiring back-and-forth interaction\footnote{For example, \cite{Jorge:etal:2016} explore agents playing a ``Guess Who'' game to learn about the emergence of question-asking and answering in language.}),  rotation of the sender and receiver roles, having a human occasionally playing one of the roles, etc.

\section{Experimental Setup}
\label{sec:experimental-setup}

\paragraph{Images} We use the McRae et al.'s
(\citeyear{McRae:etal:2005}) set of 463 base-level concrete concepts
(e.g., \emph{cat, apple, car}\ldots) spanning across 20 general
categories (e.g., \emph{animal}, \emph{fruit/vegetable},
\emph{vehicle}\ldots). We randomly sample 100 images of each concept
from ImageNet~\citep{Deng:etal:2009}. To create target/distractor
pairs, we randomly sample two concepts, one image for each concept and
whether the first or second image will serve as target. We apply to
each image a forward-pass through the pre-trained VGG
ConvNet~\citep{Simonyan:Zisserman:2014}, and represent it with the
activations from either the top 1000-D softmax layer (\emph{sm}) or the
second-to-last 4096-D fully connected layer (\emph{fc}). 
%

\begin{figure}
\includegraphics[scale=0.65]{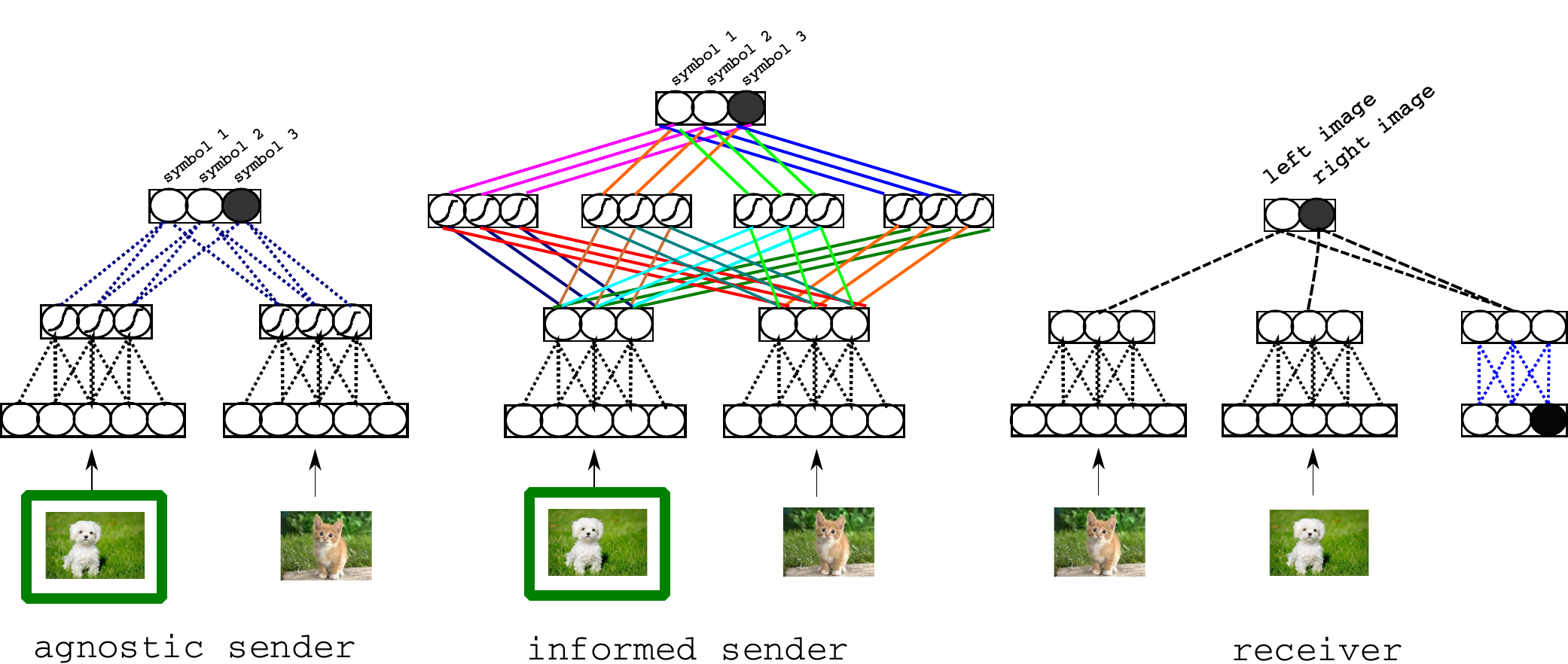}
\caption{Architectures of agent players.
}
\label{fig:players_new}
\end{figure}

\paragraph{Agent Players} Both sender and receiver are simple
feed-forward networks.  For the sender, we experiment with the two
architectures depicted in Figure \ref{fig:players_new}. Both  sender architectures take as input
the target (marked with a green square in Figure \ref{fig:players_new}) and distractor representations, always in this order, so that they are implicitly informed of which image is the target (the receiver, instead, sees the two images in random order). 

The \emph{agnostic} sender is a generic neural network that maps the
original image vectors onto a ``game-specific'' embedding space (in
the sense that the embedding is learned while playing the game)
followed by a sigmoid nonlinearity.   Fully-connected
weights are applied to the embedding concatenation to produce scores over vocabulary
symbols.

The \emph{informed} sender also first embeds
the images into a ``game-specific'' space. It then applies 1-D
convolutions (``filters'') on the image embeddings by treating them as
different channels. The informed sender uses convolutions with 
kernel size 2x1
applied dimension-by-dimension to the two image embeddings (in
Figure~\ref{fig:players_new}, there are 4 such filters). This is followed
by the sigmoid nonlinearity. The resulting feature maps are combined through
another filter (kernel size $f$x1, where $f$ is the number of filters
on the image embeddings), to produce scores for the vocabulary
symbols.  Intuitively, the informed sender has an inductive bias towards combining the two images dimension-by-dimension whereas the agnostic sender does not (though we note the agnostic architecture nests the informed one).

For both senders, motivated by the
discrete nature of language, we enforce a strong communication
bottleneck that discretizes the communication protocol. Activations on
the top (vocabulary) layer
are converted to a Gibbs distribution 
(with temperature parameter $\tau$), and then a
single symbol $s$ is sampled from the resulting probability
distribution.

The receiver takes as input the target and distractor image vectors in
random order, as well as the symbol produced by the sender (as a one-hot
vector over the vocabulary).  It embeds the images and
the symbol into its own ``game-specific'' space. It then computes dot
products between the symbol and image embeddings. Ideally, dot similarity should be higher for the image that is better denoted by the symbol. %
The two dot products are 
converted to a  Gibbs distribution 
 (with temperature $\tau$) and the
receiver ``points'' to an image by sampling from the resulting
distribution.




\paragraph{General Training Details} We set the following
hyperparameters without tuning: embedding dimensionality: 50, number of filters applied to embeddings by
  informed sender: 20, temperature of Gibbs distributions: 10. We explore two vocabulary sizes: 10 and 100 symbols. 
  
The sender and receiver parameters $\theta=\langle
\theta_{R}, \theta_{S}\rangle$ are learned while playing the
game. No weights are shared and the only supervision used is communication success, i.e.,
whether the receiver pointed at the right referent. 

This setup is naturally
modeled with Reinforcement Learning \citep{Sutton:Barto:1998}. As
outlined in Section \ref{sec:general-framework}, the sender follows policy $s(\theta_S (i_L, i_R, t))\in V$ and the receiver policy $r(i_L, i_R, s(\theta_S (i_L, i_R, t))) \in \lbrace L, R \rbrace$. The loss function that the two agents must minimize is  $-\Expect_{\tilde{r}} [R(\tilde{r})]$ where $R$ is the reward function returning 1 iff $r(i_L, i_R, s(\theta_S (i_L, i_R, t)) = t$. %
Parameters are updated through the Reinforce
rule~\citep{Williams:1992}.  We apply mini-batch updates, with a batch
size of 32 and for a total of 50k iterations (games).  At test time,
we compile a set of 10k games using the same method as for the training games.

We now turn to our main questions. The first is whether the agents can
learn to successfully coordinate in a reasonable amount of time. The second is whether the agents' language can be thought of as ``natural language'',  i.e., symbols are assigned to meanings that make intuitive sense in terms of our conceptualization of the world.

\section{Learning to Communicate}
\label{sec:exp1}

Our first question is whether agents converge to successful
communication at all. We see that they do: agents almost perfectly coordinate in the 1k rounds following the 10k training games for every architecture and parameter choice (Table \ref{tab:exp1_table}).

\begin{figure}[tb]
\centering
\includegraphics[scale=0.39]{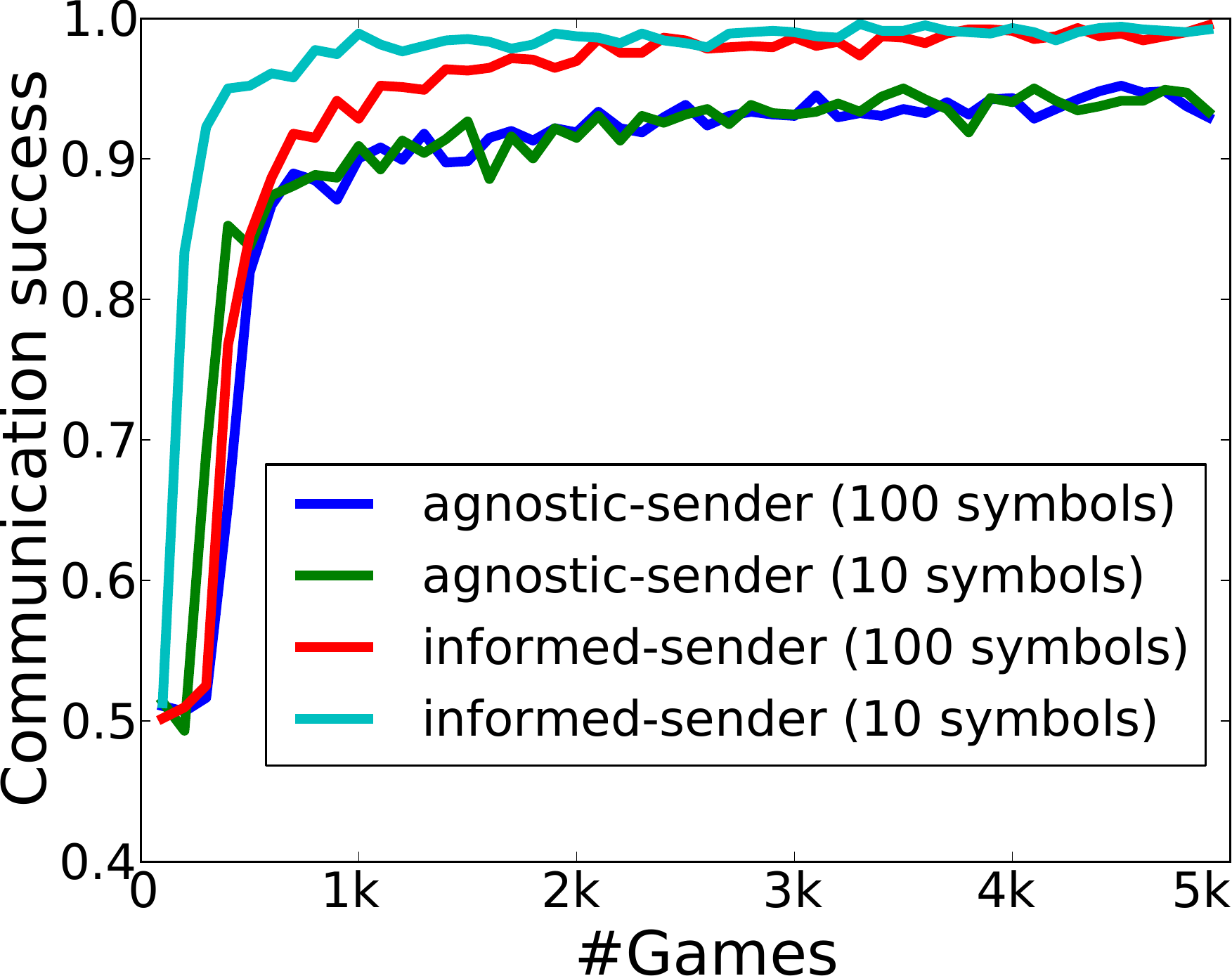}
\hspace{15pt}
\includegraphics[scale=0.34]{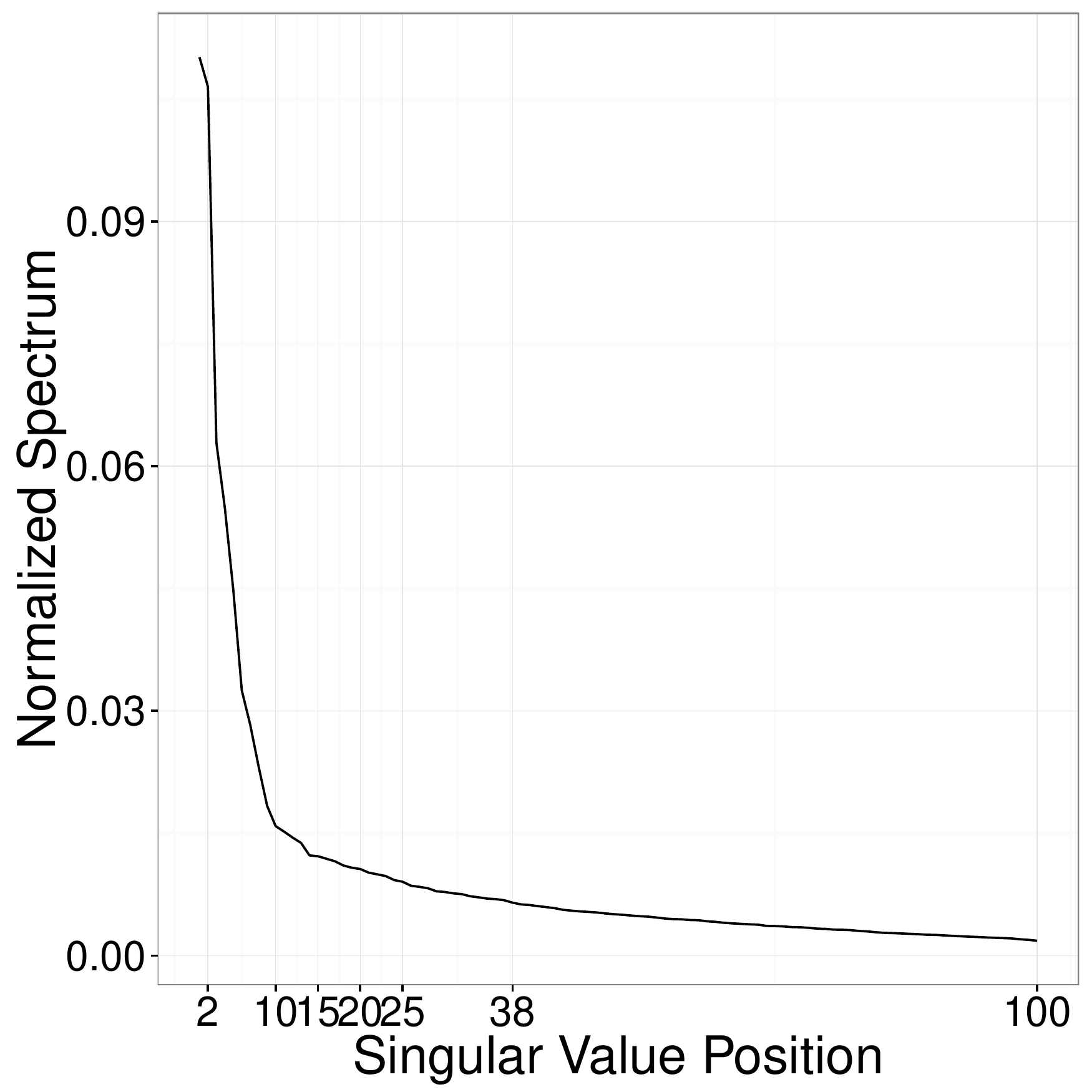}
\caption{\textbf{Left:} Communication success as a function of training iterations, we see that informed senders converge faster than agnostic ones.
 \textbf{Right:} Spectrum of an example symbol usage matrix: the first few dimensions do capture only partial variance, suggesting that the usage of more symbols by the informed sender is not just due to synonymy.}
\label{fig:exp1_comm}
\end{figure}

We see, though, some differences between different sender architectures. Figure \ref{fig:exp1_comm} (left) shows performance on a sample of the test set as a function of the first 5,000 rounds of training. The agents converge to coordination quite fast, but the informed sender reaches higher levels more quickly than the agnostic one.


\begin{table}
\centering
\small
\begin{tabular}{c|c|c|c|c|c|c|c}
id & sender   &vis   &voc &used     &comm    &      purity ($\%$)   &obs-chance\\
 	&	& rep & size& symbols & success ($\%$) &  & purity ($\%$) \\\hline
1&informed&sm     &100       &58          &100                &46           &27 \\
2&informed&fc        &100       &38          &100                &41           &23 \\
3&informed&sm     &10        &10          &100                &35           &18 \\
4&informed&fc        &10        &10          &100                &32           &17 \\
5&agnostic&sm     &100       &2           &99                 &21           &15 \\
6&agnostic&fc        &10        &2           &99                 &21           &15 \\
7&agnostic&sm     &10        &2           &99                 &20           &15 \\
8&agnostic&fc        &100       &2           &99                 &19           &15 \\
\end{tabular}

\caption{Playing the referential game: test results after 50K training games. \emph{Used symbols} column reports number of distinct vocabulary symbols that were produced at least once in the test phase. See text for explanation of \emph{comm success} and \emph{purity}. All purity values are highly significant ($p<0.001$) compared to simulated chance symbol assignment when matching observed symbol usage. 
  The \emph{obs-chance purity} column reports the difference between observed and expected purity under chance.}
  \label{tab:exp1_table}
\end{table}

The informed sender makes use of more symbols from the available vocabulary, while the agnostic
sender constantly uses a compact 2-symbol vocabulary. This suggests
that the informed sender is using more varied and word-like symbols
(recall that the images depict 463 distinct objects, so we would
expect a natural-language-endowed sender to use a wider array of
symbols to discriminate among them). However, it could also be the
case that the informed sender vocabulary simply contains higher
redundancy/synonymy. To check this, we construct a (sampled) matrix where rows are game image pairs, columns are symbols, and entries represent how often that symbol is used for that pair. We then decompose the matrix through SVD. If the sender is indeed just using a strategy with few effective symbols but high synonymy, then we should expect a $1$- or $2$-dimensional decomposition. Figure~\ref{fig:exp1_comm} (right) plots the normalized spectrum of this matrix. While there is some redundancy in the matrix (thus potentially implying there is synonymy in the usage), the language still requires multiple dimensions to summarize (cross-validated SVD suggests 50 dimensions).

We now turn to investigating the semantic properties of the emergent
communication protocol. Recall that the vocabulary that agents use is
arbitrary and has no initial meaning. One way to understand its
emerging semantics is by looking at the relationship between symbols
and the sets of images they refer to. 

The objects in our images were categorized into 20 broader
categories (such as \emph{weapon} and \emph{mammal}) by
\cite{McRae:etal:2005}. If the agents converged to higher level semantic meanings
for the symbols, we would expect that objects belonging to the same
category would activate the same symbols, e.g., that, say, when the
target images depict bayonets and guns, the sender would use the same
symbol to refer to them, whereas cows and guns should not share a
symbol. 

To quantify this, we form clusters by grouping objects by the symbols that
are most often activated when target images contain them. We then
assess the quality of the resulting clusters 
by measuring their \emph{purity} with respect to the McRae
categories. Purity \citep{Zhao:Karypis:2001} is a standard measure of
cluster ``quality''. The purity of a clustering solution is the proportion of category labels in the clusters that agree with the respective cluster majority category. This number reaches 100\% for perfect clustering and we always compare the observed purity to the score that would be obtained from a random permutation of symbol assignments to objects. Table \ref{tab:exp1_table} shows that purity, while far from perfect,
is significantly above chance in all cases. We confirm moreover that
the informed sender is producing symbols that are more semantically
natural than those of the agnostic one. 

Still, surprisingly, purity is
significantly above chance even when the latter is only using two
symbols. From our qualitative evaluations, in this case the agents converge to a
(noisy) characterization of objects as ``living-vs-non-living'' which, intriguingly, has been recognized as the most basic one in the human semantic system \citep{Caramazza:Shelton:1998}.

Rather than using hard clusters, we can also ask whether symbol usage reflects the semantics of the visual space. To do so we construct vector representations for each object (defined by its ImageNet label) by
averaging the CNN fc representations of all category images in our
data-set (see Section \ref{sec:experimental-setup} above).  Note that the fc layer, being near the top of a deep CNN, is expected to capture high-level visual properties of objects
\citep{Zeiler:Fergus:2014}. Moreover, since we average across many
specific images, our vectors should capture rather general, high-level
properties of objects. 

\begin{figure}[tb]
\centering
\includegraphics[scale=0.37]{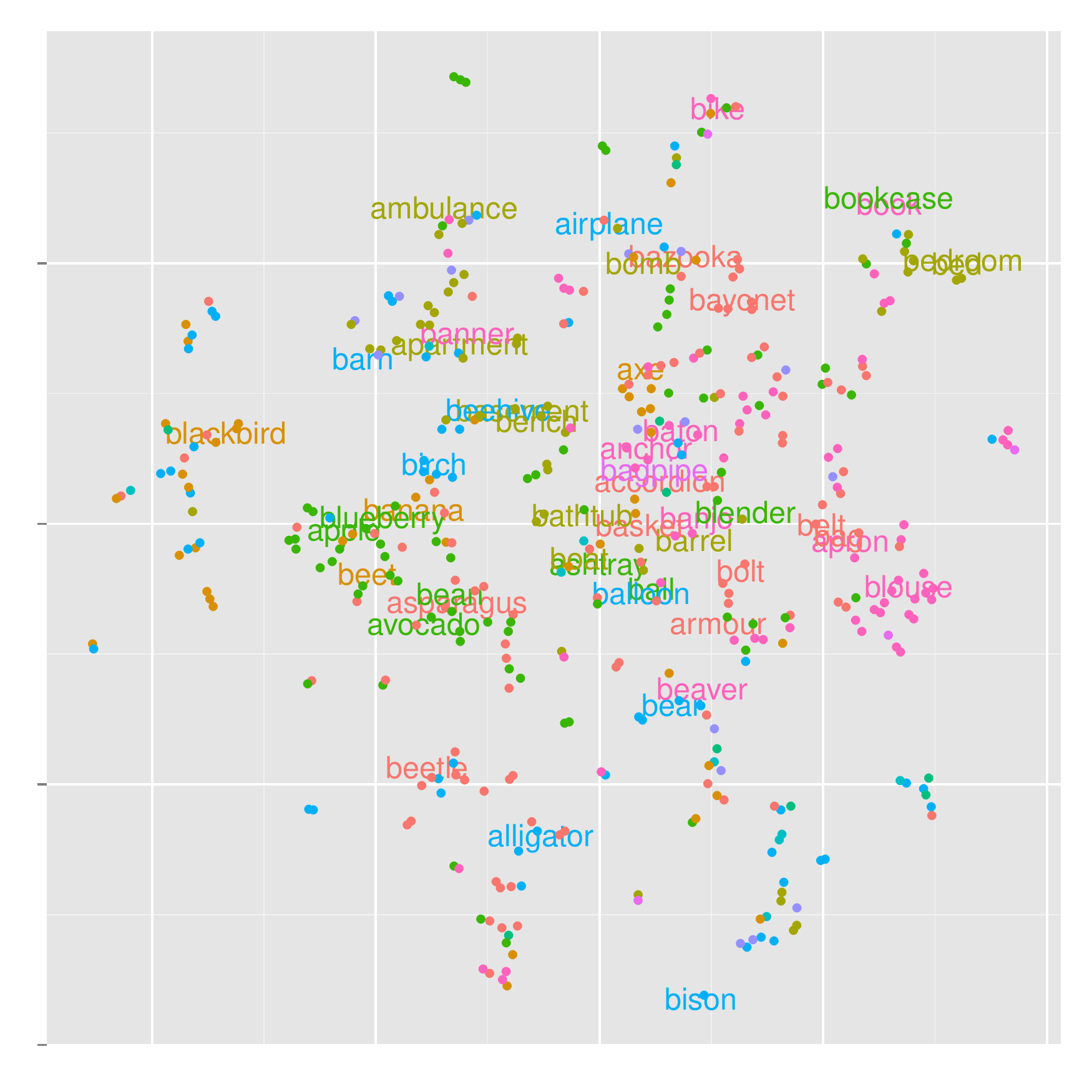}
\includegraphics[scale=0.37]{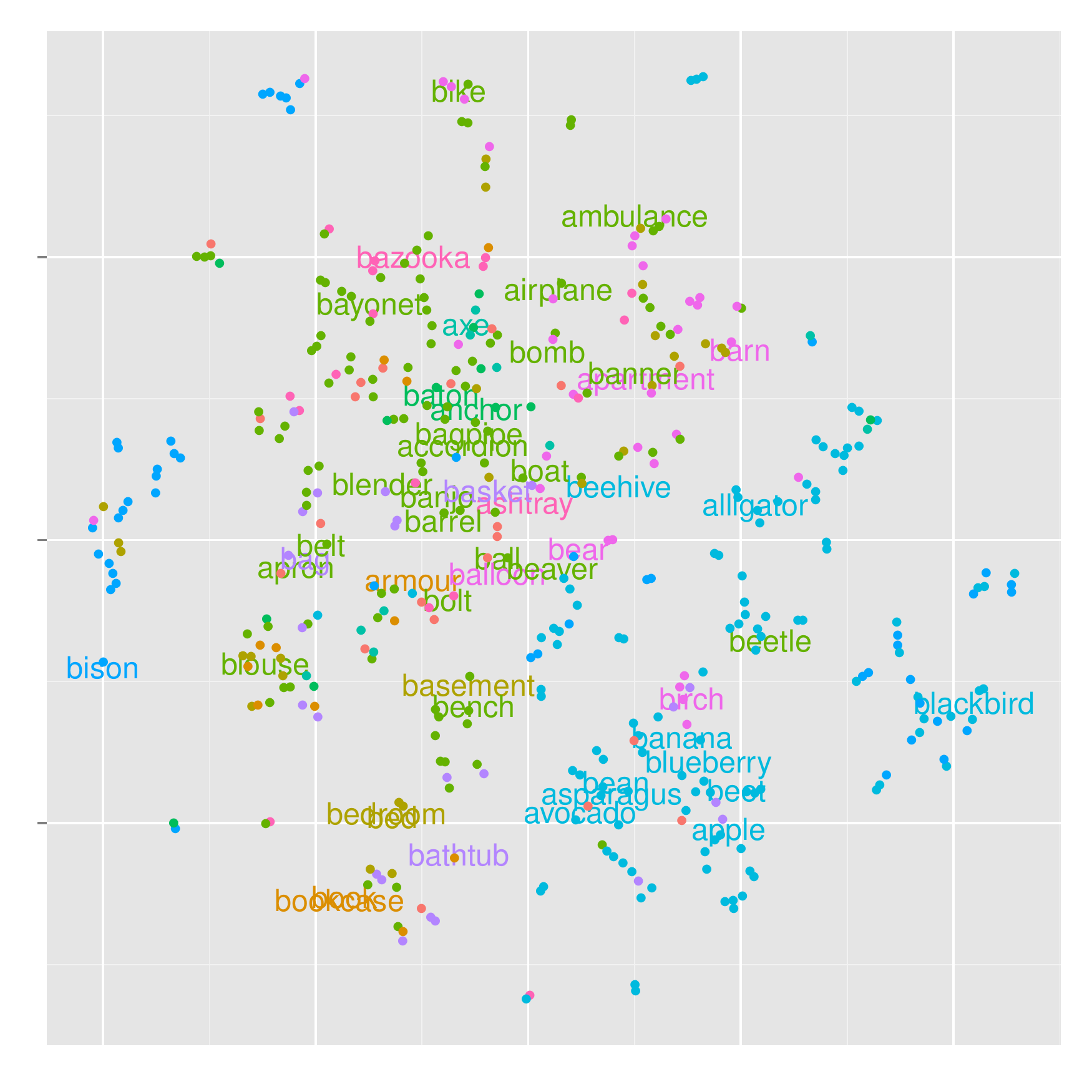}
\caption{t-SNE plots of object fc vectors color-coded by majority symbols assigned to them by informed sender. Object class names shown for a random subset. \textbf{Left:} configuration of 4th row of Table~\ref{tab:exp1_table}. \textbf{Right:} 2nd row of Table~\ref{tab:exp2_multiinstance}.}
\label{fig:exp1_tsne}
\end{figure}

We map these average object vectors to 2
dimensions via t-SNE mapping \citep{Laurens:Hinton:2008} and we
color-code them by the majority symbol the sender used for images
containing the corresponding object. Figure
\ref{fig:exp1_tsne} (left) shows the results for the current experiment. We see that objects that
are close in CNN space (thus, presumably, visually similar) are
associated to the same symbol (same color). However, there still appears to be quite a bit of variation.

\subsection{Object-level reference}
\label{sec:exp2}

We established that our agents can solve the coordination problem, and
we have at least tentative evidence that they do so by developing
symbol meanings that align with our semantic intuition. We turn now to
a simple way to tweak the game setup in order to encourage the agents
to further pursue high-level semantics. 

The strategy is to remove some
aspects of ``common knowledge'' from the game. Common knowledge, in
game-theoretic parlance, are facts that everyone knows, everyone knows
that everyone knows, and so on
\citep{brandenburger2014hierarchies}. Coordination can only occur if
the basis of the coordination is common knowledge
\citep{rubinstein1989electronic}, therefore if we remove some facts
from common knowledge, we will preclude our agents from coordinating
on them. In our case, we want to remove facts pertaining to the
details of the input images, thus forcing the agents to coordinate on
more abstract properties.
We can remove all low-level common knowledge by letting the agents
play only using class-level properties of the objects. We achieve this by modifying the game to show the agents different pairs of images but maintaining the ImageNet class of both the target and distractor (e.g., if the
target is \emph{dog}, the sender is shown a picture of a Chihuahua and
the receiver that of a Boston Terrier).


\begin{table}[tb]
\centering
\small
\begin{tabular}{c|c|c|c|c|c|c|c}
id &sender   &vis   &voc &used     &comm    &      purity ($\%$)   &obs-chance\\
 &		& rep & size& symbols & success($\%$) &  & purity ($\%$) \\\hline
1&informed&fc      &100     &43          &100               &45         &21\\
2&informed&fc      &10      &10          &100               &37         &19\\
3&agnostic&fc      &100     &2           &92               &23         &7\\
4&agnostic&fc      &10      &3           &98               &28         &12\\
\end{tabular}
\caption{Playing the referential game with image-level targets: test results after 50K training plays. Columns as in Table \ref{tab:exp1_table}. All purity values significant at $p<0.001$.}
\label{tab:exp2_multiinstance}
\end{table}

Table~\ref{tab:exp2_multiinstance} reports results for various
configurations. We see that the agents are still able to
coordinate. Moreover, we observe a small increase in symbol usage
purity, as expected since agents can now only coordinate on general
properties of object classes, rather than on the specific properties
of each image. This effect is clearer in Figure \ref{fig:exp1_tsne} (right), when we repeat t-SNE based visualization of the relationship that emerges between visual embeddings and the words used to refer to them in this new experiment.

\section{Grounding Agents' Communication in Human Language}
The results in Section~\ref{sec:exp1} show  communication robustly
arising in our game, and that we can change the environment to nudge agents 
to develop symbol meanings which are more closely related to the visual or class-based semantics of the images. %
Still, we would like agents to converge on a language fully
understandable by humans, as our ultimate goal is to develop
conversational machines. To do this, we will need to ground the
communication.%

Taking inspiration from AlphaGo \citep{Silver:etal:2016}, an AI that
reached the Go master level by combining interactive learning in games
of self-play with passive supervised learning from a large set of
human games, we combine the usual referential game, in which agents
interactively develop their communication protocol, with a supervised
image labeling task, where the sender must learn to assign objects
their conventional names. This way, the sender will naturally be
encouraged to use such names with their conventional meaning to
discriminate target images when playing the game, making communication
more transparent to humans.

In this experiment, the sender switches,
equiprobably, between game playing and a supervised image
classification task using ImageNet classes. %
Note that the supervised objective does not aim at improving agents'
coordination performance. Instead, supervision provides them with basic
grounding in natural language (in the form of image-label
associations), while concurrent interactive game playing should teach
them how to effectively use this grounding to communicate.

We use the informed sender, fc image representations and a vocabulary size of 100. Supervised training is
based on 100 labels that are a subset of the object names in our
data-set (see Section \ref{sec:experimental-setup} above). When
predicting object names, the sender  uses the usual game-embedding layer 
coupled with a softmax layer of dimensionality 100 corresponding to
the object names. Importantly, the game-embedding layers used in
object classification and the reference game are shared. Consequently,
we hope that, when playing, the sender will produce symbols aligned with object names acquired in the
supervised phase. %


The supervised objective has no negative effect on
communication success: the agents are still able to reach full
coordination after 10k training trials (corresponding to 5k trials of reference game playing). The sender uses 
many more symbols after training than in any previous experiment
(88) and symbol purity dramatically increases to 70\% (the
obs-chance purity difference also increases to 37\%). 

Even more importantly, many symbols have now become directly
interpretable, thanks to their direct correspondence to
labels. Considering the 632 image pairs where the target gold
standard label corresponds to one of the labels that were used in the
supervised phase, in 47\% of these cases the sender produced exactly
the symbol corresponding to the correct supervised label for the
target image (chance: 1\%). 


For image pairs where the target image belongs to one of the directly
supervised categories, it is not surprising that the sender adopted
the ``conventional'' supervised label to signal the target . However,
a very interesting effect of supervision is that it \emph{improves the
  interpretability of the code even when agents must communicate about
  images that do not contain objects in the supervised category
  set}. This emerged in a follow-up experiment in which, during
training, the sender was again exposed (with equal probability) to the
same supervised classification task as above, but now the agents %
played the referential game on a different dataset of images derived
from ReferItGame~\citep{Kazemzadeh:etal:2014}.  In its general format,
the ReferItGame contains annotations of bounding boxes in real images
with referring expressions produced by humans when playing the
game. 
For our purposes, we constructed 10k pairs by randomly sampling two
bounding boxes, to act as target and distractor.  Again, the agents
converged to perfect communication after 15k trials, and this time
used all 100 available symbols in some trial.

We then asked whether this language was human-interpretable. For each
symbol used by the trained sender, we randomly extracted 3 image pairs
in which the sender picked that symbol and the receiver pointed at the
right target (for two symbols, only 2 pairs matched these criteria,
leading to a set of 298 image pairs).  We annotated each pair with the
word corresponding to the symbol in the supervised set. Out of the 298
pairs, only 25 (8\%) included one of the 100 words among the
corresponding referring expressions in ReferItGame. So, in the large
majority of cases, the sender had been faced with a pair not
(saliently) containing the categories used in the supervised phase of
its training, and it had to produce a word that could, at best, only
indirectly refer to what is depicted in the target image. %
We then tested whether this code would be understandable by humans. In essence, it is as if we replaced the trained agent receiver with a human.

\begin{figure}[tb]
\centering
\includegraphics[scale=0.4]{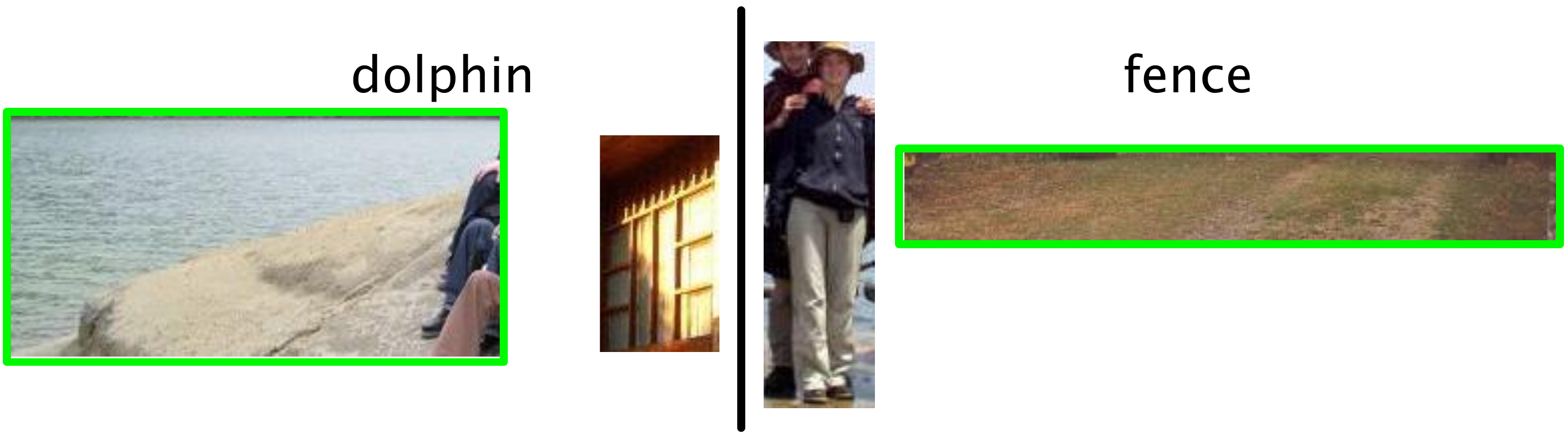}
\caption{Example pairs from the ReferItGame set, with word produced by
  sender. Target images framed in green.}
\label{fig:dolphin_fence}
\end{figure}

We prepared a crowdsourced survey using the CrowdFlower
platform. 
For each pair,
human participants were shown the two images and the sender-emitted word (that is, the ImageNet label associated to the symbol produced by the sender; see
examples in Figure \ref{fig:dolphin_fence}). The participants were
asked to pick the picture that they thought was most related to the
word. We collected 10 ratings for each pair.

We found that in 68\% of the cases the subjects were able to guess the
right image. A logistic regression predicting subject image choice from
ground-truth target images, with subjects and words as random effects, confirmed the highly significant correlation between the
true and guessed images ($z=16.75$, $p<0.0001$). Thus, while far from perfect, we find that supervised learning on a separate data set does provide some grounding for communication with humans, that generalizes beyond the conventional word denotations learned in the supervised phase. 

Looking at the results qualitatively, we found that very often sender-subject
communication succeeded when the sender established a sort of
``metonymic'' link between the words in its possession and the
contents of an image. Figure \ref{fig:dolphin_fence} shows an example where the sender produced \emph{dolphin} to refer to a picture showing a stretch of sea, and \emph{fence} for a patch of land.  Similar semantic shifts are a core
characteristic of natural language \citep[e.g.,][]{Pustejovsky:1995},
and thus subjects were, in many cases, able to successfully play the
referential game with our sender (10/10 subjects guessed the dolphin
target, and 8/10 the fence). This is very encouraging. Although the language developed in referential games will  be initially very
limited, if both
agents and humans possess the sort of flexibility displayed in this
last experiment, the
noisy but shared common ground might suffice to establish basic
communication.
\section{Discussion}

Our results confirmed that fairly simple neural-network agents can
learn to coordinate in a referential game in which they need to
communicate about a large number of real pictures. 
They also
suggest that the meanings agents come to assign to symbols in this setup
 capture general conceptual properties of the objects depicted in the
image, rather than low-level visual properties. We also showed a path to grounding the communication in natural language by mixing the game with a supervised task.

In future work, encouraged by our preliminary experiments
with object naming, we want to study how to ensure that the emergent
communication stays close to human natural language. 
Predictive learning should be retained as an
important building block of intelligent agents, focusing on teaching
them structural properties of language (e.g., lexical choice, syntax
 or style).  However, it is also important to learn the function-driven facets of language, 
such
as how to hold a conversation, and interactive games are a potentially fruitful method to achieve this goal. 

\subsubsection*{Acknowledgments}
We would like to thank Tomas Mikolov, Nghia The Pham and Sara Veldhoen for their feedback and suggestions.

\bibliography{marco,angeliki,alex}
\bibliographystyle{iclr2017_conference}

\end{document}